\documentclass[10pt,twocolumn,letterpaper]{article}

\usepackage{isba}
\usepackage{times}
\usepackage{epsfig}
\usepackage{epstopdf}
\usepackage{graphicx}
\usepackage{amsmath}
\usepackage{amssymb}
\usepackage{url,cite}
\usepackage{textcomp}

\graphicspath{{figs/}}

\isbafinalcopy 

\ifisbafinal\fi
\begin{document}

\title{Deep Feature-based Face Detection on Mobile Devices}


\author{Sayantan Sarkar$^{1}$,
 Vishal M. Patel$^{2}$ and Rama Chellappa$^1$\\
$^1$ Center for Automation Research, University of Maryland, College Park, MD 20742\\
{\tt\small \{ssarkar2, rama\}@umiacs.umd.edu}\\
$^2$ Department of Electrical and Computer Engineering, Rutgers University, Piscataway, NJ 08854\\
{\tt\small vishal.m.patel@rutgers.edu}
}

\maketitle
\thispagestyle{empty}

\begin{abstract}
   We propose a deep feature-based face detector for mobile devices to detect user's face acquired by the front-facing camera. The proposed method is able to detect faces in images containing extreme pose and illumination variations as well as partial faces.   The main challenge in developing deep feature-based algorithms for mobile devices is the constrained nature of the mobile platform and the non-availability of CUDA enabled GPUs on such devices. Our implementation takes into account the special nature of the images captured by the front-facing camera of mobile devices and exploits the GPUs present in mobile devices without CUDA-based frameworks, to meet these challenges.
   
\end{abstract}


\section{Introduction}
Current methods of authenticating users on mobile devices are mostly PIN or pattern based, which  provides authentication only during the initial login. Password-based methods are susceptible, because people sometimes set passwords that are easy to guess  or are repetitive \cite{NYTimes} and pattern-based systems are vulnerable to smudge attacks \cite{Smudge_Attack}. Once the attacker successfully bypasses the initial authentication barrier, the phone has no way of blocking or denying the attacker. Continuous authentication systems deal with this issue by continuously monitoring the user identity after the initial access to the mobile device based on how the user interacts with the mobile device. Examples of such systems include touch gesture-based systems \cite{Touchalytics}, \cite{Heng_WACV2015}, \cite{Continuous_HST2012}, face-based systems \cite{face_eye_mobile}, \cite{UMDAA}, \cite{attributes_BTAS2015}, gait-based systems \cite{MobileGait}, stylometry-based methods \cite{Lex_stylometry}, speech and face-based method \cite{mobio}\cite{mobiodatabase}  and sensor-based methods \cite{Jain_AA_ICB2015}, \cite{context_AA}. It has been shown that face-based recognition can be very effective for continuous authentication \cite{mobio}, \cite{UMDAA}, \cite{Heng_FG2015_DA}, \cite{attributes_BTAS2015}.

Face detection is a very important step in face-based authentication systems. There has been substantial progress in detecting faces in images, which have impressive performances on challenging real-world databases \cite{MSR_FD_TR_2012}. But such databases are predominantly composed of general surveillance or media type images and not specifically of images captured using front-facing cameras of smartphones. As we shall discuss later, face images captured using the front-facing cameras of mobile devices possess some unique features that can be used as powerful prior information to simplify the task of face detection on mobile platforms. This paper proposes a deep convolutional neural network (DCNN)-based face detection scheme for mobile platforms.

\subsection{Motivation} 

State of the art face detection techniques are based on DCNNs \cite{FD_BTAS2015}, \cite{multiviewFD_RCNN}.  Variations of DCNNs have been shown to perform well in various datasets like Face Detection Dataset and Benchmark  (FDDB) \cite{fddbTech} and Annotated Face in-the-Wild  (AFW) \cite{AFW_dataset_CVPR2012}. Though DCNN-based methods can run on serial processors like CPUs, they are prohibitively slow without parallel processors like GPUs. Mobile devices and consumer electronics products like cameras often have in-built face detection systems, but since they do not have much computational horsepower, simpler detection algorithms are implemented on them, which do not have as high a performance as DCNN-based methods but can run on low power mobile platforms. Thus, there is a tradeoff between high performance and hardware and power constraints.
This paper seeks to reconcile the two competing objectives and studies the feasibility and effectiveness of DCNN-based face detection methods in mobile platforms. Clearly, the most powerful DCNN-based face detectors that are designed to run on desktop environments will not be a good candidate for a DCNN-based detector for mobile platforms. Below are a few differences between the two tasks.
\begin{enumerate}
\item Differences in hardware and software setup:
\begin{itemize}
\item The de facto hardware requirement for DCNNs is a powerful Nvidia GPU. Clearly, mobile GPUs are much less powerful, hence the algorithms need to be simpler.
\item Most DCNN frameworks use a CUDA backend, but since most mobile GPUs are not made by Nvidia, they do not support CUDA. Hence, a more portable software stack is needed.
\end{itemize}

\item Differences in dataset:
\begin{itemize}
\item Generic face databases may have images with multiple small faces while the front-facing camera captures face images when the user is using the phone and hence may have one large face image. Therefore, we can restrict ourselves to detecting a single face only. Also, given the typical distance at which the user interacts with his or her phone, we can make assumptions about the maximum and minimum sizes of the captured faces.

\item The images captured by the front-facing camera usually have the user's face in a frontal pose. Extreme pose variations are rare and one can focus on detecting faces with minor pose variations.

\item Faces captured by the front-facing camera, however, tend to be partial. A mobile face detector should be equipped to detect partial faces, which is not the focus of many generic face detectors.

\end{itemize}
\end{enumerate}

This paper makes the following contributions:
\begin{itemize}
\item Exploiting the unique nature of the face detection problem on mobile platforms, we design an effective, simplified DCNN-based algorithm for mobile platforms that need not be as powerful as general face detectors, but is fine-tuned to work in a mobile setting.
\item Most of the existing implementations of DCNNs use a CUDA backend, but most mobile GPUs are not Nvidia GPUs, hence they do not support CUDA. We develop libraries (in OpenCL and RenderScript) to implement DCNN-based algorithms on GPUs without resorting to CUDA, so that the algorithm is portable across multiple platforms.
\end{itemize}

Rest of the paper is organized as follows.  We first survey related works that have influenced the current algorithm and discuss their advantages and disadvantages. Section~\ref{sec:two} introduces the algorithm in full details and ends with a discussion on the salient features of the algorithm.  Section~\ref{sec:three} explores the details of the actual implementation of the algorithm on a mobile platform.  Section~\ref{sec:four} presents evaluation results of the algorithm on two datasets, UMD-AA and MOBIO. Finally we draw some conclusions about the algorithm and suggest some future directions.

\subsection{Related Work}
Cascade classifiers form an important and influential family of face detectors. Viola-Jones detector \cite{Viola_Jones} is a classic method, which provides realtime face detection, but works best for full, frontal, and well lit faces. Extending the work of cascade classifiers, some authors \cite{relwork1} have trained multiple models to address pose variations. An extensive survey of such methods can be found in \cite{MSR_FD_TR_2012}.

Modeling of the face by parts is another popular approach. Zhu \emph{et al.} \cite{AFW_dataset_CVPR2012} proposed a deformable parts model that detected faces by identifying face parts and modeling the whole face as a collection of face parts joined together using \textquotesingle springs\textquotesingle . The springs like constraints were useful in modeling deformations, hence this method is somewhat robust to pose and expression changes.

 \begin{figure*}[t]
	\begin{center}
		   \includegraphics[width=.83\linewidth]{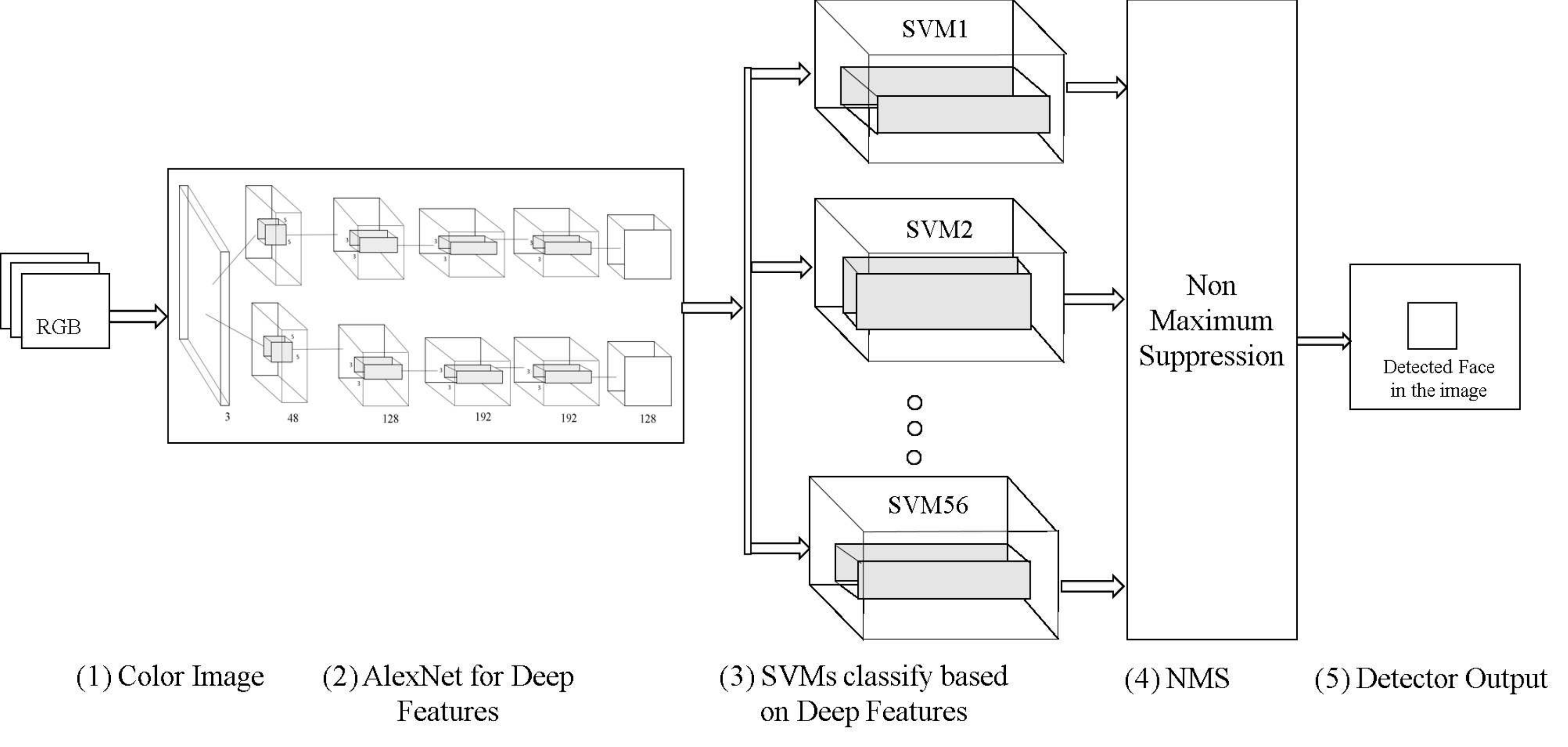}
	\end{center}
	   \caption{Overview of the proposed deep feature-based face detection algorithm for mobile devices.}
		 \label{fig:blockdiag}
\end{figure*}
As mentioned before, current state-of-the-art methods involve deep networks, which have been extensively adopted and studied both by the academic community and industry. Current face detectors at commercial companies like Google and Facebook use massive datasets to train very deep and complex networks that work well on unconstrained datasets, but they require huge training datasets and powerful hardware to run.

Recent studies have shown that in the absence of massive datasets or hardware infrastructure, transfer learning can be effective as it allows one to introduce deep networks without having to train it from scratch. This is possible as lower layers of deep networks can be viewed as feature extractors, while higher layers can be tuned to the task at hand. Therefore, one can use the lower layers of common deep networks like AlexNet \cite{Krizhevsky_imagenetclassification} to extract general features, that can then be used to train other classifiers. Works of Bengio \emph{et al.} \cite{YosinskiCBL14} have studied how transfer learning works for deep networks.

Specific to the mobile platform, Hadid \emph{et al.} \cite{face_eye_mobile} have demonstrated a local binary pattern (LBP)-based method on a Nokia N90 phone. Though it is fast, it is not a robust method and was designed for an older phone. Current phones have more powerful CPUs and more importantly, even GPUs, which can implement DCNNs.

Finally, let us consider the datasets used for mobile face detection. While there are many face databases available, they are not suitable for evaluating mobile face detection algorithms. MOBIO is a publicly available mobile dataset \cite{mobio} which consists of bi-modal (audio and video) data taken from 152 people, but it is a very constrained one as users are asked to keep their faces within a certain region, so that full faces are captured. A more suitable dataset for our purpose is the semi-constrained UMD-AA dataset \cite{UMDAA}, which shall be described in a later section.

\section{Deep Features-based Face Detection on Mobile Devices}
\label{sec:two}
As mentioned briefly before, transfer learning is an effective way to incorporate the performance of deep networks. The first step of the Deep Features based Face Detection on Mobiles (DFFDM) algorithm is to extract deep features using the first 5 layers of Alexnet. Different sized sliding windows are considered, to account for faces of different sizes and an SVM is trained for each window size to detect faces of that particular size. Then, detections from all the SVMs are pooled together and some candidates are suppressed based on an overlap criteria. Finally, a single bounding box is output by the detector. In the following subsections, the details of the algorithm and model training are provided. Figure~\ref{fig:blockdiag} provides an overview of the entire system.

\subsection{Dataset}
The UMD-AA dataset is a database of 720p videos and touch gestures of users that are captured when the user performs some given tasks on a mobile device (iPhone) \cite{UMDAA}. There are 50 users (43 males and 7 females) in the database, who perform 5 given tasks (eg, typical tasks like scrolling, reading, viewing images etc.) in three illumination conditions (a room with natural light, a well-lit room and a poorly lit room). A total of 8036 images, spread over all users and all sessions, were extracted from these video recordings and manually annotated with bounding boxes for faces. Of these 6429 images had user\textquotesingle s faces in the frame and 1607 were without faces, or with faces at extreme poses, with eyes and nose not visible or a very small partial face visible in the frame, which are all the cases when we can safely say there is no face present in the frame.

\subsection{Training SVMs}
\begin{figure}[t]
	\begin{center}
		   \includegraphics[width=.8\linewidth]{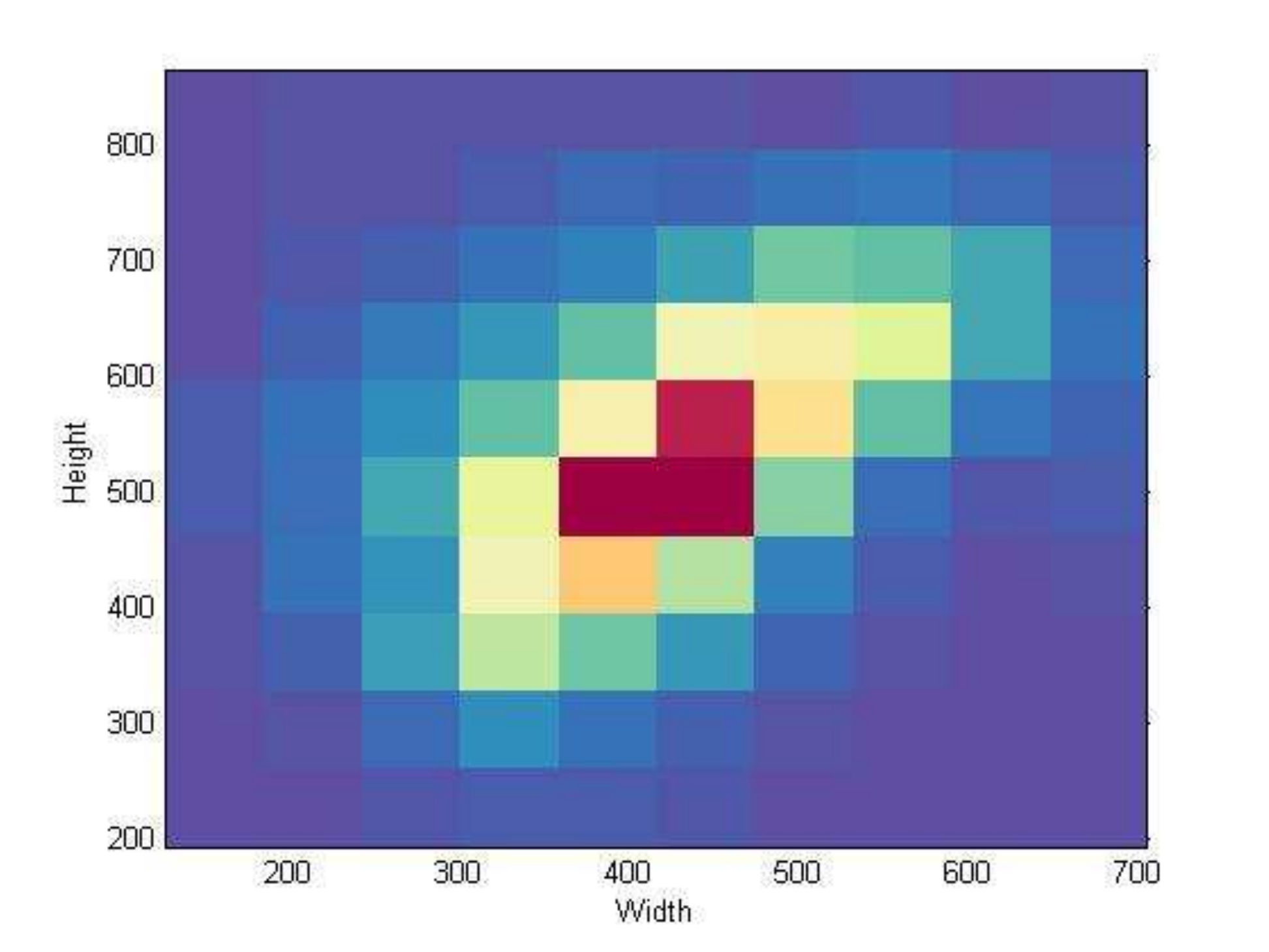}
	\end{center}
	   \caption{A histogram showing distribution of bounding box widths and heights.}
		 \label{fig:hist}
\end{figure}

For training, 5202 images from the UMD-AA database is used. Analysing the distribution of face sizes, we find that the height of faces vary from around 350 to 700 and the width varies from 300 to 600. A 2D histogram of the height and widths of the faces in the dataset are shown in Figure~\ref{fig:hist}.

Now the images are captured at 720p resolution (1280 rows x 720 columns). But since that resolution is too high for our purpose, we resize it to 640 x 360. Therefore typical faces range from 175 to 350 rows and 150 to 300 columns in this reduced resolution.

First we extract deep features from these resized images by forwarding them through AlexNet \cite{Krizhevsky_imagenetclassification}. We tap the network at the 5th convolutional layer (after max-pooling). The standard AlexNet reduces an image by a factor of 16 in both dimensions. Thus, if the $k^{\text{th}}$ input image is of size $p_{k} \times q_{k}$, the output is of dimensions $w_{k} \times h_{k} \times 256$, where the feature space width $w_{k}$ and height $h_{k}$ are given by (\ref{eq0})

\begin{equation} \label{eq0}
w_{k} = \left \lceil{p_{k}/16}\right \rceil , h_{k} = \left \lceil{q_{k}/16}\right \rceil. 
\end{equation}

The 3rd dimension is 256 because the $5^{\text{th}}$ layer of AlexNet uses 256 filters. Given the typical face dimensions in the last paragraph, they are reduced by a factor of 16 in the feature space to heights ranging from 10 to 22 and widths ranging from 9 to 19 approximately. Obviously, a single sized sliding window cannot account for these varying sizes, therefore we consider windows of width starting from 8 and increasing to 20 in steps of 2, and height starting from 9 and increasing in steps of 2 to 23. In total we get 56 different window sizes for which we need to train 56 different SVMs. We denote a window by $W_{ij}$, where $i$ denotes its window height and $j$ denotes its window width. 

Let $w_{k}$ and $h_{k}$, as defined in (\ref{eq0}), denote the width and height of the deep feature for the face in the $k^{\text{th}}$ training image. The face from the $k^{\text{th}}$ training image is used as a positive sample for the SVM $W_{ij}$, if Eq. (\ref{eq1}) is satisfied.
\begin{equation} \label{eq1}
|{i-h_{k}}| \le t_{p} ~  \& ~ |{j-w_{k}}| \le t_{p},
\end{equation}
for some threshold for selecting positive samples, $t_p$. That is, we select those faces for $W_{ij}$ whose sizes are comparable and close to the window\textquotesingle s dimensions. 

For negative samples, we extract random patches of size $i\times j$ from those training samples which have no faces. If the $k^{\text{th}}$ training sample has a face of size $w_{k} \times h_{k}$, and for a particular window $W_{ij}$, if (\ref{eq2}) holds,
\begin{equation} \label{eq2}
|{i-h_{k}}| > t_{n} ~  \& ~ |{j-w_{k}}| > t_{n},
\end{equation}
for some threshold for selecting negative samples, $t_n$, then we extract a few random patches from the $k^{\text{th}}$ training sample that act as negative samples for $W_{ij}$. That is, if the face in an image is of a very different size from the current window $W_{ij}$ under consideration, we extract negative samples from it, so that $W_{ij}$ gives a negative response of faces of different size. Finally, since the UMD-AA database does not have many images with no faces, we extract some random negative patches from images of the UPenn Natural Image Database \cite{10.1371/journal.pone.0020409}.

Once we have extracted the positive and negative samples for each window size, we discard those window sizes which do not have enough positive examples. Then we convert the three dimensional deep feature patches into a single dimensional feature vector. Thus for $W_{ij}$, we get a feature vector of length $i \times j \times 256$. We estimate the mean and standard deviation of features from each window, which are used to normalize the features.

Next we train linear SVMs for each window. Since we get a very long feature vector, it is difficult to train an SVM with all positive and negative samples together. To make the training tractable, we divide the samples into batches and train over many cycles. Specifically, let $p_{ij}$ be the number of positive samples for $W_{ij}$. Then we choose a small number of negative samples say $n_{ij}$ and train the SVM. Then we find the scores of the $n_{ij}$ negative training samples using the weights we get after training and retain only those that are close to the separating hyperplane and discard the rest. We refill the negative samples batch with new negative samples and continue this cycle multiple times. This procedure is performed for each SVM.

\subsection{Full Face Detection Pipeline}
After the SVMs are trained, we can scan the deep feature extracted from the given image $k$ in a sliding window fashion for each SVM. Specifically for an image of size $w_{k} \times h_{k}$, the deep feature is of $h_{k}$ rows and $w_{k}$ columns as given by (\ref{eq0}) and 256 depth. Therefore, for $W_{ij}$, we can slide the window from position $(1, 1)$, which is the top left, to $(h_{k}-i, w_{k}-j)$. Let $(r_{ij}, c_{ij})$ denote the position where the SVM yields highest score. Then we say that a bounding box, whose top left is at $16 \times (r_{ij}, c_{ij})$ and has width $16 \times j$ and height $16 \times i$ is the prediction from $W_{ij}$. Note that we multiply by 16, because the feature space\textquotesingle s height and width is approximately 16 times smaller than that of the original image.

Now that we have 1 prediction from each of the 56 SVMs, we need to combine them to get a single prediction. A modified version of the non maximal suppression scheme used by Girshick \emph{et al.}  \cite{girshick2014rich} is used for this purpose. First we sort the 56 proposals by their scores and then pick the candidate with the highest score. Boxes that overlap significantly with it and have a considerably lower score than it are ignored. This is continued for the next highest scoring candidate in the list, till all boxes are checked. After this we process the remaining candidates by size. If a larger box significantly overlaps a smaller box, but the larger box has a slightly lower score than the smaller box, we suppress the smaller box. This is useful in the following scenario: A smaller SVM may give a strong response for part of a full face, while the larger SVM responsible for detecting faces of that size may give a slightly lower response. But clearly the larger SVM is making the correct prediction, so we need to suppress the overlapping smaller SVM\textquotesingle s candidate. After performing these suppressions, we pick the SVM\textquotesingle s candidate that has the highest score. We then choose a suitable threshold, and if final candidate\textquotesingle s score is larger than that, we declare a face is present at that location, else declare that there is no face present.

\subsection{Salient Features}
Sliding window approaches usually work on the principle of extracting appropriate features and then sliding a window and deciding if an object is present in that window or not. The proposed algorithm, DFFDM, can be thought of as using DCNNs to extract the features for the sliding window approach. However, to make the sliding window approach work for detecting faces of varying scales, we need to extract features across scaled versions of the input image. The approach followed by Ranjan \emph{et al.} in \cite{FD_BTAS2015} is based on extracting deep features at multiple resolutions of the image and then training a single SVM to detect faces. 

Clearly extracting deep features is a very costly operation because of the sheer number of convolutions involved. Passing the image at multiple resolutions through the network increases the workload even more. Therefore, the proposed algorithm passes the image through the DCNN only once, but trains SVMs of different sizes to achieve scale invariance. Also, the different SVM sizes help in detecting partial faces. For example, tall and thin windowed SVMs are usually trained with left half or right half faces, while short and fat windowed SVMs are trained for top half of faces. SVMs whose aspect ratio match a normal full face's aspect ratio are trained on full faces. Thus, different sized windows help in scale invariance as well as in detecting partial faces.

\section{Implementation}
\label{sec:three}
Current popular deep learning platforms include Caffe, Theano and Torch. Although, these platforms have a CPU only version, they are significantly slower than the GPU enabled versions. These platforms have a CUDA based backend that offloads the heavy, but parallelizable, computations involved in a convolutional deep network to an Nvidia GPU. Nvidia has been actively developing and supporting deep learning research and has released optimized libraries such as cuDNN. Thus, although there are multiple frameworks in the deep learning system, the computational backend is dominated by CUDA based-code and Nvidia GPUs.

Unfortunately, CUDA is proprietary and works only for Nvidia's CUDA enabled GPUs. Therefore, existing deep learning frameworks are difficult to port on to GPUs made by other vendors. Current mobile devices have GPUs that are predominantly provided by Adreno, Mali and PowerVR. Nvidia's mobile processor Tegra does power some phones and tablets, and these devices support CUDA, but the overwhelming majority of devices do not have CUDA enabled GPUs.

OpenCL \cite{Stone:2010:OPP:622179.1803953} is an open standard, developed by Khronos Group, to support multiple vendors and facilitate cross platform heterogeneous and parallel computing. All major vendors like Qualcomm, Samsung, Apple, Nvidia, Intel and ARM conform to the OpenCL standard. Thus OpenCL is a portable option for implementing convolutional networks in GPUs other than those made by Nvidia. Recently though, Google has developed RenderScript to facilitate heterogeneous computing on the Android platform.

Mobile devices are obviously not an ideal platform to perform training on massive datasets. But once the model has been trained, we can hope to run the forward pass on mobile platforms. Thus to harness GPUs of mobile devices to perform the convolution heavy forward pass, we have implemented OpenCL and RenderScript-based libraries. The OpenCL library is general and should work on any GPU, while the RenderScript library is specifically tailored for Android. An Android specific example is the use of Schraudolp's fast exponentiation \cite{Schraudolph99} to approximately but quickly compute the normalization layer in AlexNet. Full exponentiation takes a significant amount of time and can become bottlenecks in weaker mobile GPUs.

The OpenCL and RenderScript libraries implement the primary ingredients for a basic convolutional deep network: convolution and activation layers, max pooling layers and normalization layers, each of which can be parallelized on GPUs. By appropriately stacking up these layers in the correct combination and initializing the network with pre-trained weights we can build a CNN easily. For our purpose we have implemented the AlexNet network as described earlier, but we can easily build other networks given its weights and parameters. For an image of size 360x640, a single forward pass, running on a machine with 4th generation Intel Core i7 and Nvidia GeForce GTX 850M GPU, takes about 1 second for the OpenCL implementation. For an image of the same size, on the Renderscript implementation running on different phones, we summarize the run time results in Table \ref{tab:runtime}. Only about 10\% or less of this run time is due to max-pooling layer, normalization layer, SVMs and non maximum suppression. The rest of the time is due to the heavy computations of the convolutional layers. Continuously running the algorithm on a Nexus 5 drains the battery at 0.45\% per minute, while leaving the phone undisturbed drains the battery at around 0.16\% per minute.

\begin{table}[htp!]
\centering
\begin{tabular}[htp!]{ | p{1.6cm} |  p{1.2cm} | p{1.2cm} |p{2.4cm}|}
\hline
Phone & Runtime & GPU & CPU \\ \hline \hline
Moto G &  36.7 s & Adreno 305 & Qualcomm Snapdragon 400\\ \hline
HTC One (M7) &  31.2 s & Adreno 320 & Qualcomm Snapdragon 600\\ \hline
Samsung Galaxy S4 &  28.0 s & Adreno 320 & Qualcomm Snapdragon 600\\ \hline
Nexus 5   & 11.9 s & Adreno 330 & Qualcomm Snapdragon 800 \\ \hline
LG G3   & 10.3 s &  Adreno 330 & Qualcomm Snapdragon 801 \\ \hline
Nexus 6   & 5.7 s &  Adreno 420 & Qualcomm Snapdragon 805 \\ \hline
\end{tabular}
\caption{Run times of DFFDM on different mobile platforms}
\label{tab:runtime}
\end{table}

\section{Evaluation and Results}
\label{sec:four}
For evaluation, we consider common metrics like Precision-Recall plots, F1 scores and Accuracy. We compare the performance of our algorithm on the UMD-AA \cite{UMDAA} and MOBIO \cite{mobiodatabase}\cite{mobio} datasets with Deep Pyramid Deformable Part Model (DP2MFD) \cite{FD_BTAS2015}, which is among the state-of-the-art algorithms for some challenging datasets like AFW and FDDB, deformable part model for face detection (DPM) \cite{AFW_dataset_CVPR2012} and Viola Jones detector (VJ) \cite{Viola_Jones}. 

We compute detections based on 50\% intersection over union criteria. Let $d$ be the detected bounding box, $g$ be the ground truth box and $s$ be the associated score of the detected box $d$. Then for declaring a detection to be valid, we need Eq. (\ref{iou}) to be satisfied for some threshold $t$

\begin{equation} \label{iou}
\dfrac{area(d\cap g)}{area(d\cup g)} > 0.5~ \& ~ s \ge t.
\end{equation}

\subsection{UMD-AA Dataset}
Results on UMD-AA dataset are summarized in Table~\ref{tab:feature}.

\begin{table}[htp!]
\centering
\begin{tabular}[htp!]{ | p{1.4cm} |  p{1.2cm} | p{1.2cm} | p{1.2cm} | p{1.2cm}|}
\hline
Metric & DFFDM & DP2MFD & DPM & VJ\\ \hline \hline
Max F1 &  92.8\% & 89.0\% &  84.1\% &  67.7\% \\ \hline
Max Accuracy   & 88.0\% & 82.3\% &  76.4\% &  58.0\% \\ \hline
Recall at 95\% precision   & 85.7\% &  81.7\% &  72.6\% &  - \\ \hline
\end{tabular}
\caption{Comparision of different metrics for various detectors on UMD-AA database}
\label{tab:feature}
\end{table}

     \begin{figure}[t]
	\begin{center}
		   \includegraphics[width=1\linewidth]{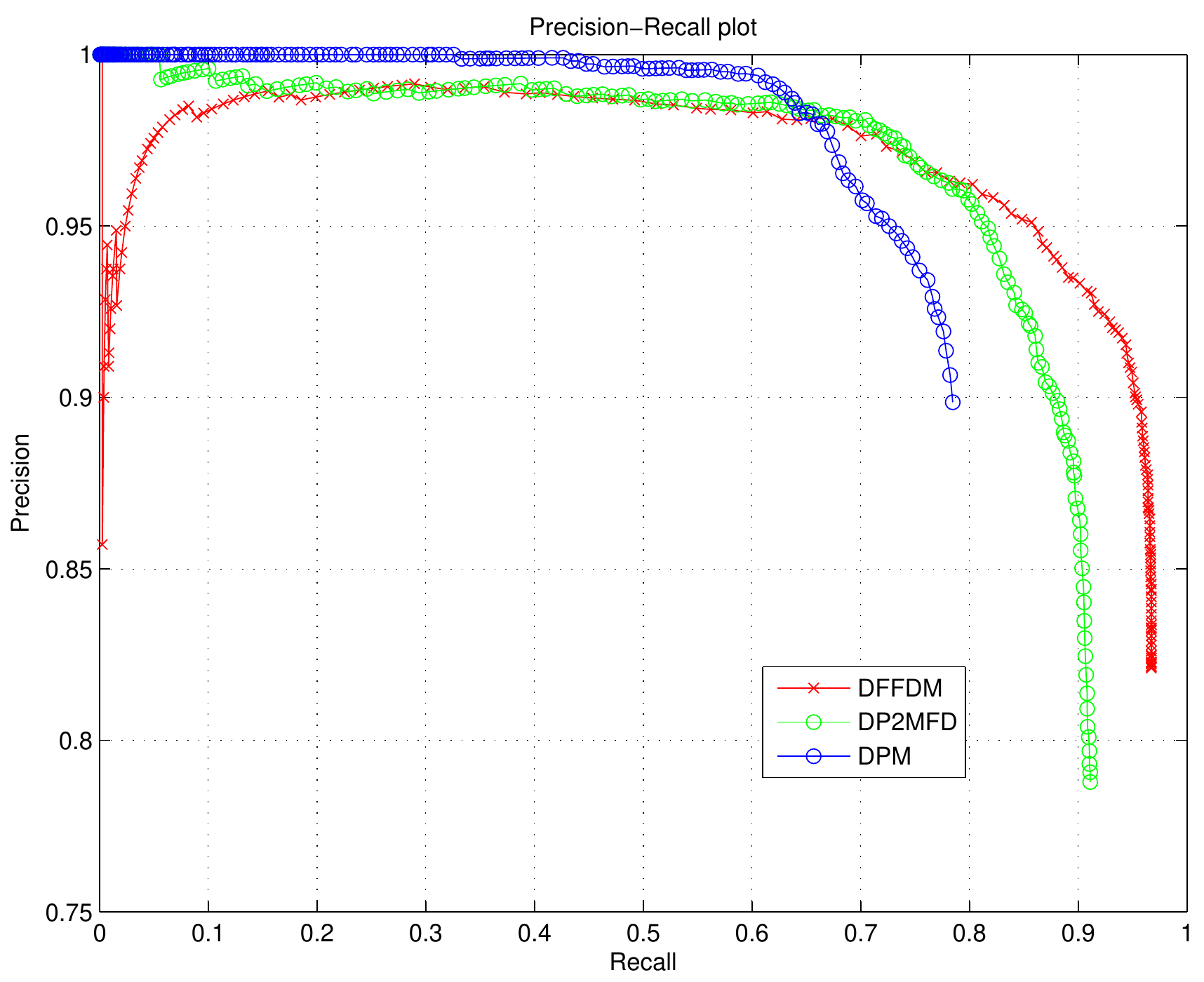}
	\end{center}
	   \caption{Precision Recall plot corresponding to the UMD-AA dataset.}
		 \label{fig:precrecall}
\end{figure}

\begin{figure}[t]
	\begin{center}
		   \includegraphics[width=.9\linewidth]{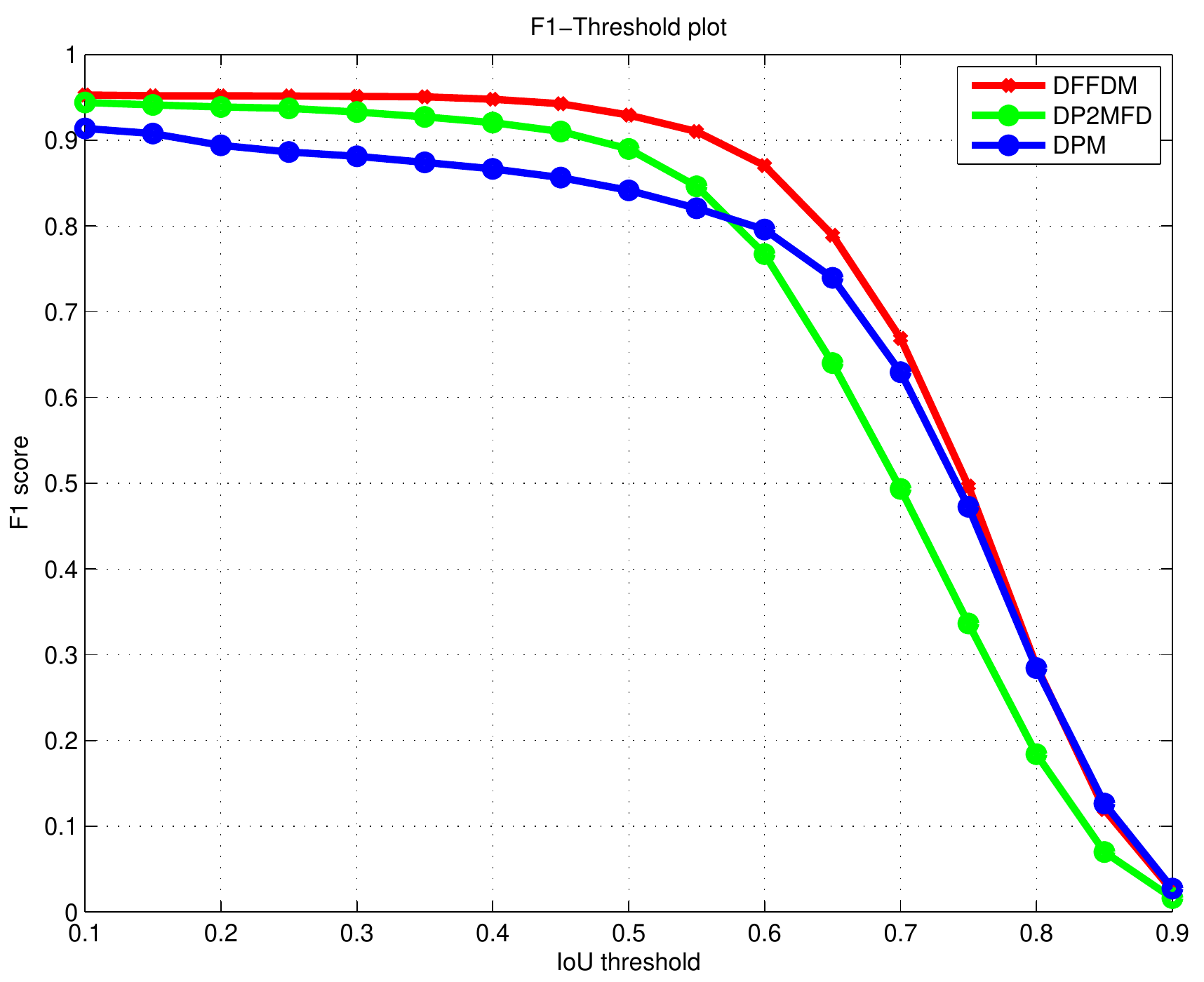}
	\end{center}
	   \caption{Plot showing variation of F1 score with respect to overlap threshold corresponding to the UMD-AA dataset.}
		 \label{fig:f1vsth}
\end{figure}

\begin{figure}[t]
	\begin{center}
		   \includegraphics[width=.9\linewidth]{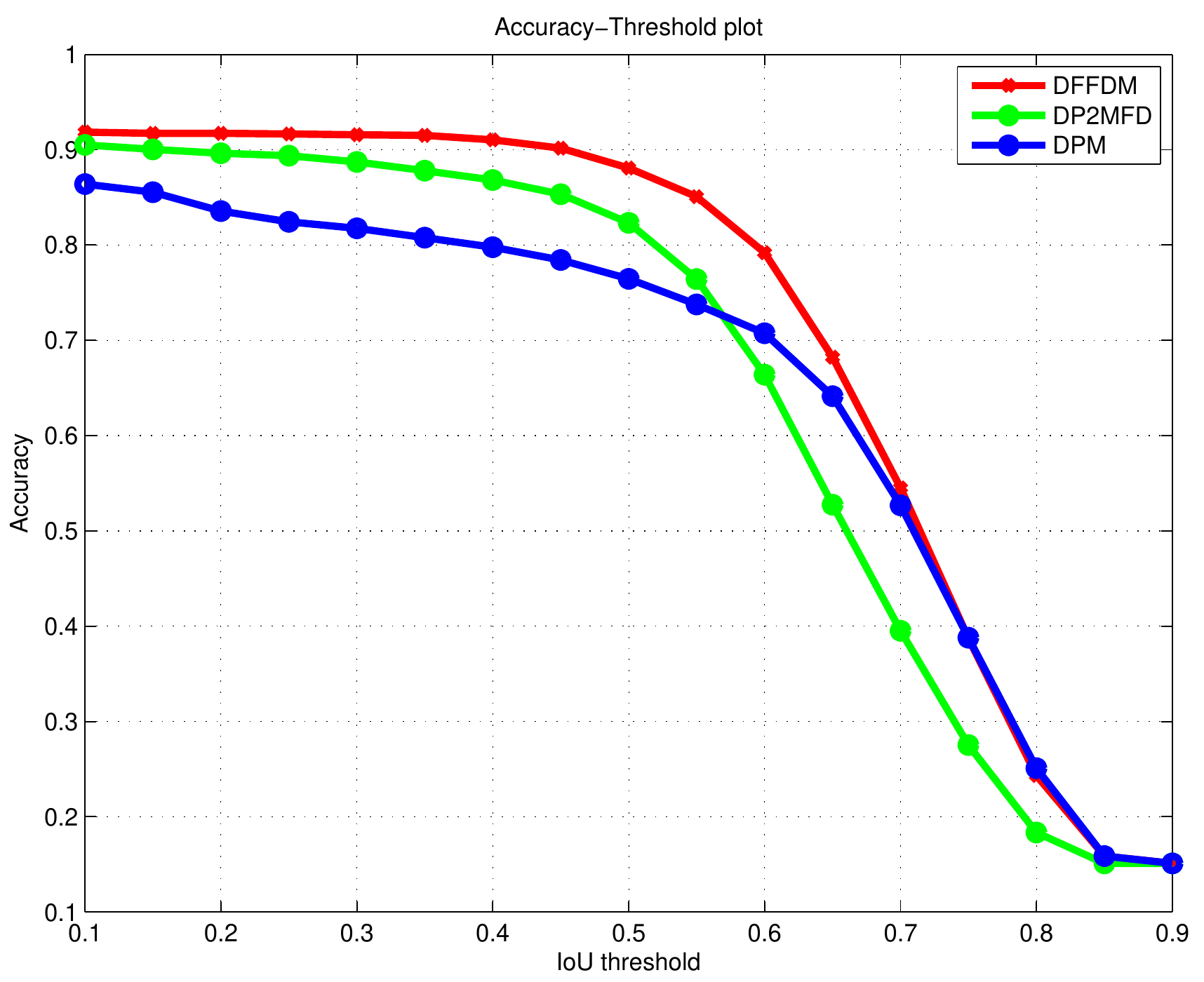}
	\end{center}
	   \caption{Plot showing variation of accuracy with respect to overlap threshold corresponding to the UMD-AA dataset.}
		 \label{fig:accvsth}
\end{figure}

To check the robustness of the detector, we vary the intersection-over-union threshold as defined in Eq. (\ref{iou}) from 0.1 to 0.9 and plot the resulting F1 score in Figure~\ref{fig:f1vsth} and accuracy in Figure~\ref{fig:accvsth}. We see that the DFFDM algorithm gives better performance at higher overlap thresholds too.

A few example positive and negative detections are shown in Figure~\ref{fig:examples}. The detections are marked in red, while the ground truth is in yellow. The first row shows a few examples of positive detections with partial faces and the second row shows positive detections with pose variations. The third row shows some false detections, or detections with score lesser than 1. The detector is quite robust to illumination change and is able to detect partial or extremely posed faces.

     \begin{figure}[t]
	\begin{center}
		   \includegraphics[width=.9\linewidth]{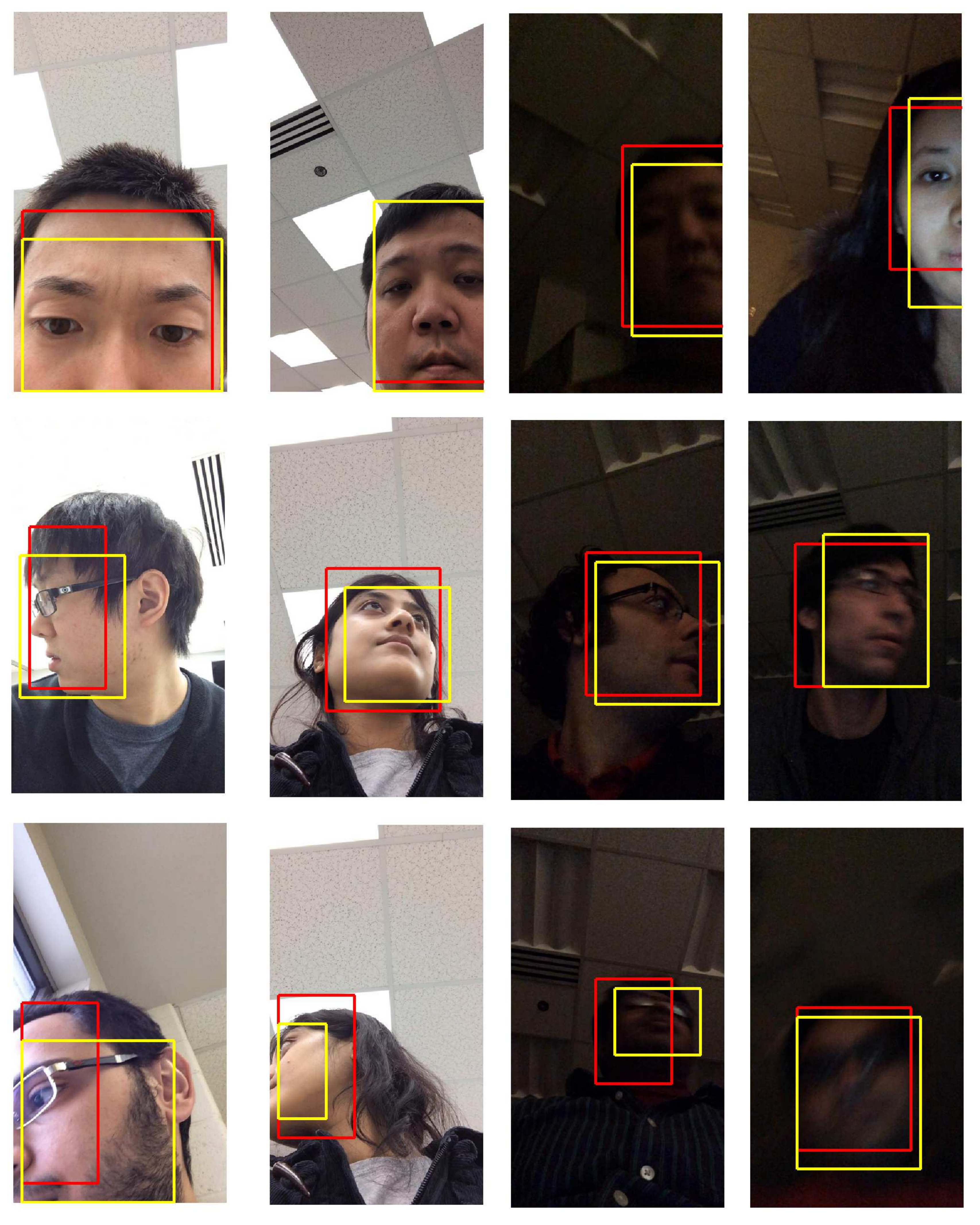}
	\end{center}
	   \caption{Examples of positive detections (with pose variations and occlusion, in first 2 rows) and examples of negative detections (due to insufficient overlap or low score in 3rd row) on UMD-AA. The detector\textquotesingle s output is in red, while ground truth is in yellow.}
		 \label{fig:examples}
\end{figure}

\subsection{MOBIO Dataset}
Results on MOBIO dataset are summarized in Table~\ref{tab:mobio}. The MOBIO dataset has full frontal faces only, therefore we get very high performance. DP2MFD beats our algorithm for this dataset, which can be attributed to the fact that DP2MFD is one of the best algorithms, trained on a large, varied dataset, and for full frontal faces it has near perfect performance over multiple scales. For DFFDM, SVMs of different sizes were trained, based on the typical size of faces captured by the front camera. But sometimes for very large or small faces, the training dataset of UMD-AA may not have enough samples, therefore for extremely scaled faces, DFFMD may fail. This can be remedied by training on a larger database, and also by training SVMs on more scales. A few example positive and negative detections are shown in Figure ~\ref{fig:examples}. The first row shows positive detections while the second row shows failures. As the examples show, there are some false detectiosn for really large faces, of which we did not have many examples in the UMD-AA training dataset on which DFFDM was trained.

\begin{table}[htp!]
\centering
\begin{tabular}[htp!]{ | p{1.4cm} |  p{1.2cm} | p{1.2cm} | p{1.2cm} | p{1.2cm}|}
\hline
Metric & DFFDM & DP2MFD & DPM & VJ\\ \hline \hline
Max F1 &  97.9\% & 99.7\% &  97.8\% &  92.6\% \\ \hline
Max Accuracy   & 96.0\% & 99.3\% &  95.8\% & 86.3\% \\ \hline
\end{tabular}
\caption{Comparision of different metrics for various detectors on MOBIO database}
\label{tab:mobio}
\end{table}

     \begin{figure}[t]
	\begin{center}
		   \includegraphics[width=1\linewidth]{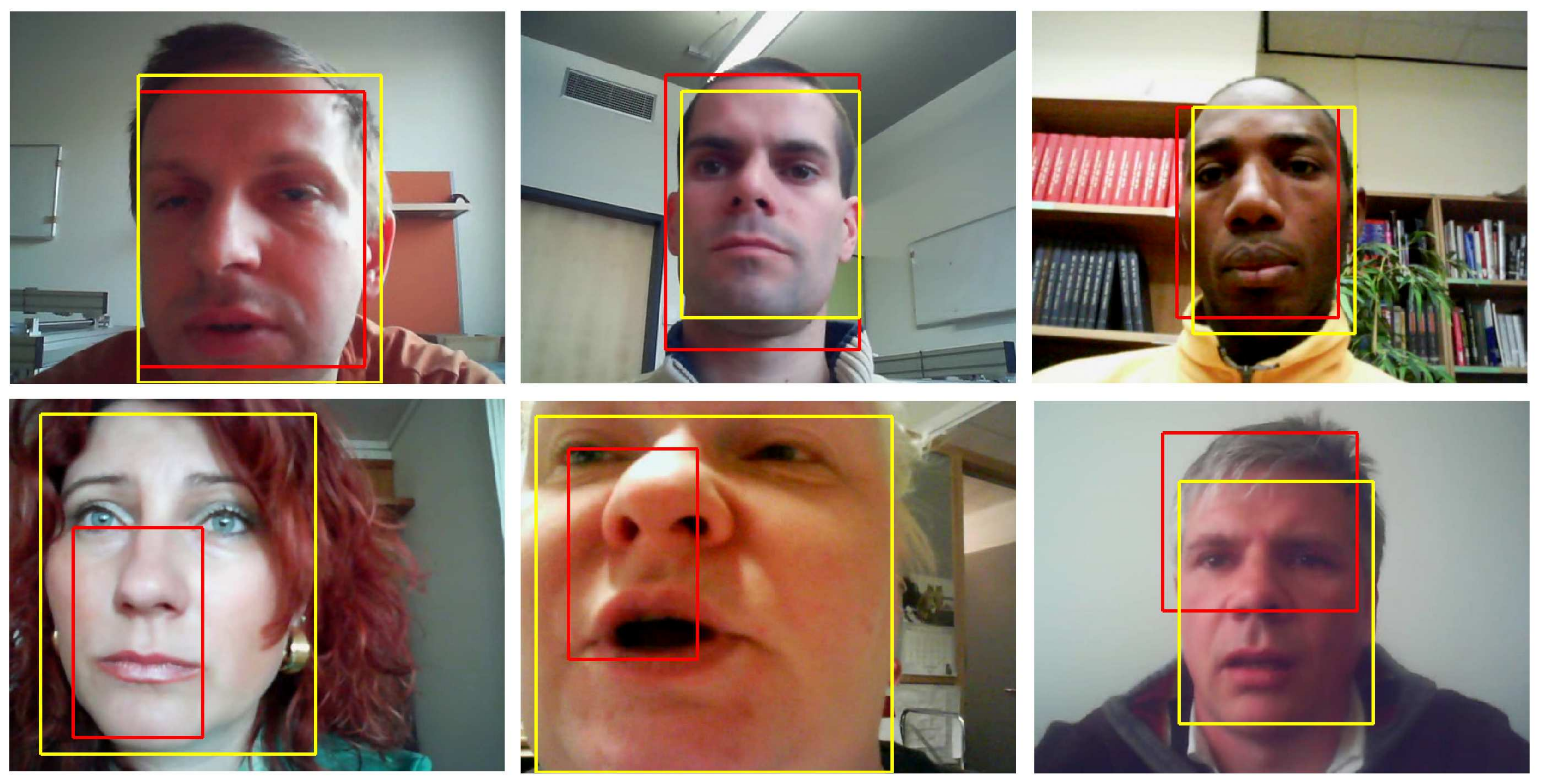}
	\end{center}
	   \caption{Examples of positive (1st row) and  negative (2nd row) detections on MOBIO. The detector\textquotesingle s output is in red, while ground truth is in yellow.}
		 \label{fig:examples}
\end{figure}

\section{Conclusion and Future Directions}
\label{sec:five}
This paper presents a deep feature based face detector for locating faces in images taken by a mobile device\textquotesingle s front camera. Keeping the constrained nature of the problem in mind, the algorithm performs only one forward pass per image and shifts the burden of achieving scale invariance to the multiple SVMs of different sizes. As is expected from DCNN-based algorithms, it outperforms traditional feature-based schemes at the cost of a longer run time. Thus although DCNN based methods do not seem suitable for real time monitoring due to their run times on mobile devices, they can still be used as a backup in case a simpler detector fails. However there is much scope of optimizations and also mobile hardware has been getting more and more powerful, which looks promising.

This study also produced OpenCL and RenderScript based libraries for implementing DCNNs, that are more portable and suitable for mobile devices than CUDA based frameworks currently in popular use.

Future directions of inquiry includes code optimizations to make the GPU utilization faster thus speeding up the whole process. Also, we wish to explore simpler DCNNs that may be more suited to the mobile environment than a full blown AlexNet. Finally, the libraries used for this algorithm are more portable than CUDA based libraries and we hope to expand on them to facilitate research on deep networks on mobile GPUs.

\section*{Acknowledgement}
This work was supported by cooperative agreement FA8750-13-2-0279 from DARPA.

{\small
\bibliographystyle{IEEEtran}
\bibliography{ISBA_FD_v3}

\begin{thebibliography}{10}
\providecommand{\url}[1]{#1}
\csname url@samestyle\endcsname
\providecommand{\newblock}{\relax}
\providecommand{\bibinfo}[2]{#2}
\providecommand{\BIBentrySTDinterwordspacing}{\spaceskip=0pt\relax}
\providecommand{\BIBentryALTinterwordstretchfactor}{4}
\providecommand{\BIBentryALTinterwordspacing}{\spaceskip=\fontdimen2\font plus
\BIBentryALTinterwordstretchfactor\fontdimen3\font minus
  \fontdimen4\font\relax}
\providecommand{\BIBforeignlanguage}[2]{{%
\expandafter\ifx\csname l@#1\endcsname\relax
\typeout{** WARNING: IEEEtran.bst: No hyphenation pattern has been}%
\typeout{** loaded for the language `#1'. Using the pattern for}%
\typeout{** the default language instead.}%
\else
\language=\csname l@#1\endcsname
\fi
#2}}
\providecommand{\BIBdecl}{\relax}
\BIBdecl

\bibitem{NYTimes}
\BIBentryALTinterwordspacing
A.~Vance. (2010, Jan) If your password is 123456, just make it hackme. [Online;
  posted JAN. 20, 2010]. [Online]. Available: \url{http://www.nytimes.com}
\BIBentrySTDinterwordspacing

\bibitem{Smudge_Attack}
A.~J. Aviv, K.~Gibson, E.~Mossop, M.~Blaze, and J.~M. Smith, ``Smudge attacks
  on smartphone touch screens,'' in \emph{Proceedings of the 4th USENIX
  Conference on Offensive Technologies}, 2010, pp. 1--7.

\bibitem{Touchalytics}
M.~Frank, R.~Biedert, E.~Ma, I.~Martinovic, and D.~Song, ``Touchalytics: On the
  applicability of touchscreen input as a behavioral biometric for continuous
  authentication,'' \emph{IEEE Transactions on Information Forensics and
  Security}, vol.~8, no.~1, pp. 136--148, Jan 2013.

\bibitem{Heng_WACV2015}
H.~Zhang, V.~M. Patel, M.~E. Fathy, and R.~Chellappa, ``Touch gesture-based
  active user authentication using dictionaries,'' in \emph{IEEE Winter
  conference on Applications of Computer Vision}.\hskip 1em plus 0.5em minus
  0.4em\relax IEEE, 2015.

\bibitem{Continuous_HST2012}
T.~Feng, Z.~Liu, K.-A. Kwon, W.~Shi, B.~Carbunar, Y.~Jiang, and N.~Nguyen,
  ``Continuous mobile authentication using touchscreen gestures,'' in
  \emph{IEEE Conference on Technologies for Homeland Security}, Nov 2012, pp.
  451--456.

\bibitem{face_eye_mobile}
A.~Hadid, J.~Heikkila, O.~Silven, and M.~Pietikainen, ``Face and eye detection
  for person authentication in mobile phones,'' in \emph{ACM/IEEE International
  Conference on Distributed Smart Cameras}, Sept 2007, pp. 101--108.

\bibitem{UMDAA}
M.~E. Fathy, V.~M. Patel, and R.~Chellappa, ``Face-based active authentication
  on mobile devices,'' in \emph{IEEE International Conference on Acoustics,
  Speech and Signal Processing}, 2015.

\bibitem{attributes_BTAS2015}
P.~Samangouei, V.~M. Patel, and R.~Chellappa, ``Attribute-based continuous user
  authentication on mobile devices,'' in \emph{International Conference on
  Biometrics Theory, Applications and Systems}, 2015.

\bibitem{MobileGait}
M.~Derawi, C.~Nickel, P.~Bours, and C.~Busch, ``Unobtrusive user-authentication
  on mobile phones using biometric gait recognition,'' in \emph{International
  Conference on Intelligent Information Hiding and Multimedia Signal
  Processing}, Oct 2010, pp. 306--311.

\bibitem{Lex_stylometry}
L.~Fridman, S.~Weber, R.~Greenstadt, and M.~Kam, ``Active authentication on
  mobile devices via stylometry, gps location, web browsing behavior, and
  application usage patterns,'' \emph{IEEE Systems Journal}, 2015.

\bibitem{mobio}
C.~McCool, S.~Marcel, A.~Hadid, M.~Pietikainen, P.~Matejka, J.~Cernocky,
  N.~Poh, J.~Kittler, A.~Larcher, C.~Levy, D.~Matrouf, J.-F. Bonastre,
  P.~Tresadern, and T.~Cootes, ``Bi-modal person recognition on a mobile phone:
  using mobile phone data,'' in \emph{IEEE ICME Workshop on Hot Topics in
  Mobile Multimedia}, Jul. 2012.

\bibitem{mobiodatabase}
C.~McCool and S.~Marcel, ``Mobio database for the icpr 2010 face and speech
  competition,'' Idiap, Idiap-Com Idiap-Com-02-2009, 11 2009.

\bibitem{Jain_AA_ICB2015}
D.~Crouse, H.~Han, D.~Chandra, B.~Barbello, and A.~K. Jain, ``Continuous
  authentication of mobile user: Fusion of face image and inertial measurement
  unit data,'' in \emph{International Conference on Biometrics}, 2015.

\bibitem{context_AA}
A.~Primo, V.~Phoha, R.~Kumar, and A.~Serwadda, ``Context-aware active
  authentication using smartphone accelerometer measurements,'' in
  \emph{Computer Vision and Pattern Recognition Workshops (CVPRW), 2014 IEEE
  Conference on}, June 2014, pp. 98--105.

\bibitem{Heng_FG2015_DA}
H.~Zhang, V.~M. Patel, S.~Shekhar, and R.~Chellappa, ``Domain adaptive sparse
  representation-based classification,'' in \emph{IEEE International Conference
  on Automatic Face and Gesture Recognition}.\hskip 1em plus 0.5em minus
  0.4em\relax IEEE, 2015.

\bibitem{MSR_FD_TR_2012}
C.~Zhang and Z.~Zhang, ``A survey of recent advances in face detection,''
  Microsoft Research, Tech. Rep. MSR-TR-2010-66, 2010.

\bibitem{FD_BTAS2015}
R.~Ranjan, V.~M. Patel, and R.~Chellappa, ``A deep pyramid deformable part
  model for face detection,'' in \emph{International Conference on Biometrics
  Theory, Applications and Systems}, 2015.

\bibitem{multiviewFD_RCNN}
S.~S. Farfade, M.~Saberian, and L.-J. Li, ``Multi-view face detection using
  deep convolutional neural networks,'' in \emph{International Conference on
  Multimedia Retrieval}, 2015.

\bibitem{fddbTech}
V.~Jain and E.~Learned-Miller, ``Fddb: A benchmark for face detection in
  unconstrained settings,'' University of Massachusetts, Amherst, Tech. Rep.
  UM-CS-2010-009, 2010.

\bibitem{AFW_dataset_CVPR2012}
X.~Zhu and D.~Ramanan, ``Face detection, pose estimation, and landmark
  localization in the wild,'' in \emph{IEEE Conference on Computer Vision and
  Pattern Recognition}, June 2012, pp. 2879--2886.

\bibitem{Viola_Jones}
P.~A. Viola and M.~J. Jones, ``Robust real-time face detection,''
  \emph{International Journal of Computer Vision}, vol.~57, no.~2, pp.
  137--154, 2004.

\bibitem{relwork1}
C.~Huang, H.~Ai, Y.~Li, and S.~Lao, ``High-performance rotation invariant
  multiview face detection,'' \emph{IEEE Transactions on Pattern Analysis and
  Machine Intelligence}, vol.~29, no.~4, pp. 671--686, 2007.

\bibitem{Krizhevsky_imagenetclassification}
A.~Krizhevsky, I.~Sutskever, and G.~E. Hinton, ``Imagenet classification with
  deep convolutional neural networks,'' in \emph{Advances in Neural Information
  Processing Systems}, p. 2012.

\bibitem{YosinskiCBL14}
\BIBentryALTinterwordspacing
J.~Yosinski, J.~Clune, Y.~Bengio, and H.~Lipson, ``How transferable are
  features in deep neural networks?'' \emph{CoRR}, vol. abs/1411.1792, 2014.
  [Online]. Available: \url{http://arxiv.org/abs/1411.1792}
\BIBentrySTDinterwordspacing

\bibitem{10.1371/journal.pone.0020409}
\BIBentryALTinterwordspacing
G.~Tkačik, P.~Garrigan, C.~Ratliff, G.~Milčinski, J.~M. Klein, L.~H.
  Seyfarth, P.~Sterling, D.~H. Brainard, and V.~Balasubramanian, ``Natural
  images from the birthplace of the human eye,'' \emph{PLoS ONE}, vol.~6,
  no.~6, p. e20409, 06 2011. [Online]. Available:
  \url{http://dx.doi.org/10.1371%2Fjournal.pone.0020409}
\BIBentrySTDinterwordspacing

\bibitem{girshick2014rich}
R.~Girshick, J.~Donahue, T.~Darrell, and J.~Malik, ``Rich feature hierarchies
  for accurate object detection and semantic segmentation,'' in \emph{Computer
  Vision and Pattern Recognition (CVPR), 2014 IEEE Conference on}.\hskip 1em
  plus 0.5em minus 0.4em\relax IEEE, 2014, pp. 580--587.

\bibitem{Stone:2010:OPP:622179.1803953}
\BIBentryALTinterwordspacing
J.~E. Stone, D.~Gohara, and G.~Shi, ``Opencl: A parallel programming standard
  for heterogeneous computing systems,'' \emph{IEEE Des. Test}, vol.~12, no.~3,
  pp. 66--73, May 2010. [Online]. Available:
  \url{http://dx.doi.org/10.1109/MCSE.2010.69}
\BIBentrySTDinterwordspacing

\bibitem{Schraudolph99}
N.~N. Schraudolph, ``A fast, compact approximation of the exponential
  function,'' 1999.

\end{thebibliography}
}



\end{document}